\documentclass[journal]{IEEEtran}
\ifCLASSINFOpdf
\else
\fi
\usepackage{xcolor}
\usepackage{color}
\definecolor{mygray}{RGB}{193,203,215}
\usepackage{booktabs}
\usepackage{multirow}
\usepackage{colortbl}
\usepackage{graphicx}
\usepackage{caption}
\DeclareCaptionFont{blue}{\color{blue}}

\usepackage{array}
\usepackage{dblfloatfix}
\usepackage{balance}
\usepackage{enumitem}
\usepackage{url}
\usepackage{dingbat}
\newcolumntype{C}[1]{>{\centering\arraybackslash}m{#1}}

\usepackage[framemethod=tikz]{mdframed}

\usepackage{bigstrut}
\usepackage{multirow}
\usepackage{amsmath}

\makeatletter
\long\def\@IEEEtitleabstractindextextbox#1{\parbox{0.922\textwidth}{#1}}
\makeatother

\usepackage[rawfloats=true]{floatrow} 
\restylefloat{figure} 
\restylefloat{table} 

\usepackage{cite}
\usepackage{amsmath,amssymb,amsfonts}
\usepackage{algorithmic}
\usepackage{caption,graphicx}
\usepackage{textcomp}
\usepackage{xcolor}
\usepackage{mathrsfs}
\usepackage{pgfplots}

\usepackage{amssymb}

\usepackage{subcaption}
\usepackage[table,xcdraw]{xcolor}
\usepackage{url}
\usepackage{textcomp}
\usepackage{xcolor}
\usepackage{mathrsfs}
\usepackage{graphicx}
\usepackage{subcaption}
\usepackage{tikz}
\usepackage{dblfloatfix}
\usepackage{bigstrut}
\usepackage{multirow}
\usepackage{color}
\usepackage{fancyhdr}
\usepackage{listings} 
\usepackage{changepage}
\usepackage{lipsum}
\usepackage{balance}
\usepackage{enumitem}
\usepackage{comment}
\usepackage{breqn}
\usepackage{dblfloatfix}
\usepackage{amsthm}
 \usepackage{makecell}

\providecommand{\cmark}{\checkmark}
\providecommand{\xmark}{\texttimes}
 
\usepackage{tikz}
\usetikzlibrary{arrows.meta,positioning,fit,shapes.multipart,calc}
\tikzset{
  box/.style={draw, rounded corners, align=center, inner sep=3pt, minimum height=10mm},
  small/.style={box, font=\scriptsize},
  arr/.style={-{Latex[length=2mm]}, line width=0.7pt},
  opt/.style={dash pattern=on 2pt off 1.5pt},
  note/.style={font=\scriptsize, align=center},
  group/.style={draw, rounded corners, inner sep=4pt}
}
\begin{document}
%
\title{R-FLoRA: Residual-Statistic-Gated Low-Rank Adaptation for Single-Image Face Morphing Attack Detection}
%
%
%

\author{Raghavendra Ramachandra,~\IEEEmembership{Senior Member,~IEEE,}
\thanks{Raghavendra Ramachandra is with the SAFE Center, Department
of Information Security and Communication Technology, Norwegian University of Science and Technology (NTNU), Norway,
 e-mail: (raghavendra.ramachandra @ ntnu.no.}
\thanks{}}

%
%

\markboth{Accepted in IEEE Transactions on Information Forensics and Security (TIFS), 2026}%
{Shell \MakeLowercase{\textit{et al.}}: Bare Demo of IEEEtran.cls for IEEE Journals}
%



\maketitle

\begin{abstract}
Face morphing attacks pose a substantial risk to the reliability of face recognition systems used in passport issuance, border control, and digital identity verification. Detecting morphing attacks from a single facial image  remains challenging owing to the lack of a trusted reference and the diversity of attack generation methods. This paper presents a new Single-Image Face Morphing Attack Detection (S-MAD) framework that integrates high-frequency Laplacian residual statistics with representations from a frozen, foundation-scale vision transformer. The approach employs residual-statistic-gated low-rank adapters (R-FLoRA) and feature-wise residual fusion (Res-FiLM) to enhance sensitivity to local morphing artefacts while preserving the semantic context of the backbone. A novel residual-contrastive alignment loss further regularises the fused token space, improving discrimination under unseen morphing conditions. Comprehensive experiments on four ICAO-compliant datasets, encompassing seven morph generation techniques, demonstrate that the proposed method consistently surpasses nine recent state-of-the-art S-MAD algorithms in detection accuracy and cross-domain (or dataset) generalisation. With a frozen backbone and minimal trainable parameters, the model achieves real-time efficiency and interpretability, making it suitable for real-life scenarios in biometric verification systems.
\end{abstract}


%
\IEEEpeerreviewmaketitle

\section{Introduction}
Face Recognition Systems (FRS) are extensively deployed in border control and identity verification applications, where their ability to establish the correspondence between a traveller's live image and a stored biometric template underpins secure and efficient processing~\cite{Venkatesh-MADSurvey-IEEETTS-2021}. Despite their widespread adoption, such systems remain vulnerable to sophisticated presentation attacks, which can compromise both the accuracy and security of automated border management~\cite{Venkatesh-MADSurvey-IEEETTS-2021}.

Among these, face morphing constitutes a serious threat to both human operators and automated facial recognition algorithms~\cite{Venkatesh-MADSurvey-IEEETTS-2021}. Face morphing is a form of image synthesis in which facial images of two or more individuals are combined to create a new, visually plausible identity~\cite{Venkatesh-MADSurvey-IEEETTS-2021}. This manipulated image can reliably match multiple subjects, allowing adversaries to circumvent border control systems and gain illegitimate access or travel credentials~\cite{Venkatesh-MADSurvey-IEEETTS-2021}. Recent studies have demonstrated that both humans and state-of-the-art FRS are highly susceptible to such attacks, underlining the urgent need for dedicated morphing attack detection methods and comprehensive countermeasures~\cite{Makrushin-Human-Vs-Algorithm-2019, Makrushin-simulation-border-control-experiment-2020, godage2022analyzing}.

Morphing Attack Detection (MAD) methods are principally classified into Differential Morphing Attack Detection (D-MAD) and Single Morphing Attack Detection (S-MAD) paradigms~\cite{Venkatesh-MADSurvey-IEEETTS-2021}. D-MAD involves the comparison of a live capture with a reference image (typically stored in the passport document chip), detecting authentication attempts when notable discrepancies indicative of morphing are observed. In contrast, S-MAD algorithms operate solely on a single image, without the benefit of an accompanying reference, making them well-suited for scenarios where only the presented image is available, as commonly arises in several border control and identity enrolment workflows~\cite{Venkatesh-MADSurvey-IEEETTS-2021}. Although D-MAD can exploit cross-sample correspondences, S-MAD presents a significantly greater technical challenge due to the limited contextual information, it is critical for real-world border control deployments where reference samples may be absent or inaccessible~\cite{Venkatesh-MADSurvey-IEEETTS-2021}. These factors motivated us to consider the S-MAD. 

Early attempts on S-MAD were predominantly centred around hand-crafted feature representations designed to capture the subtle texture distortions introduced by the morphing process~\cite{Raghavendra-DetectingMorphedFace-BTAS-2016}. Progress in the MAD techniques has led to the development of a diverse set of approaches: deep learning-based~\cite{Raghavendra-DNNMorphingDetection-CVPR-2017, Damer-anomalyMAD-BTAS-2019, Damer-PW-MAD-VisualComputing-2021}, multimodal feature-based~\cite{raghavendra2022multimodality}, unsupervised~\cite{fang2022unsupervised}, residual noise analysis methods~\cite{Venkatesh-CANnetworkMAD-WACV-2020, ResNoise_BenchMark} and texture based approaches \cite{Scherhag-MorphingAttacks-MorphingTechniques-BIOSIG-2017, Raghavendra-DetectingMorphedFace-BTAS-2016}. More recently, foundation models have become prominent in the morphing attack detection landscape~\cite{MADCLIP, ozgur2025foundpad}.

Deep learning-based S-MAD frameworks leverage both off-the-shelf pre-trained models~\cite{Raghavendra-DNNMorphingDetection-CVPR-2017} and end-to-end architectures~\cite{Damer-PW-MAD-VisualComputing-2021} to exploit hierarchical facial representations, advancing detection robustness and accuracy. In parallel, multimodal strategies have concentrated on combining cues from facial areas that are particularly prone to morphing artefacts, thus increasing reliability~\cite{raghavendra2022multimodality}. Unsupervised approaches, in contrast, typically employ anomaly detection protocols, where models are trained on bona fide samples only and then used to flag morphing attacks as outliers~\cite{fang2022unsupervised}. Residual noise analysis methods exploit image denoising or artefact extraction (for example, via diffusion models, CAN, BM3D, wavelets, and CNN denoisers) to uncover holistic noise signatures introduced by morphing; among these, CANs have often yielded the most reliable detection scores~\cite{ResNoise_BenchMark,Venkatesh-CANnetworkMAD-WACV-2020, ColorNoiseCVIP-2025}.

Recent S-MAD reviews~\cite{Venkatesh-MADSurvey-IEEETTS-2021, MAD_Survey2025} provide a comprehensive overview of these methodological avenues. Notably, ensemble approaches fusing diverse feature sets and classifier outputs such as post-processing feature fusion~\cite{PostProcessMorph} have also emerged, but may incur considerable computational expense. Efforts to address the challenge of generalisation to unseen morphing types have produced unsupervised frameworks based on self-paced and auto-encoding methods~\cite{fang2022unsupervised} and deep generative models including diffusion models trained on bona fide samples, which highlight morphs via their anomalous reconstruction error~\cite{ivanovska2023mad_ddpm}. The AlphaNet framework~\cite{tapia2023alphanet} applies MobileNetV3 to learn from morphs crafted from multiple contributors, exploiting alpha-channel filters in a multiclass setup. Extensive experiments confirm AlphaNet's competitiveness within and across datasets, with particular attention to local morphing artefact suppression. In~\cite{tapia2025single}, the authors propose a few-shot learning model using a Siamese triplet network. This architecture is optimised to maximise inter-class separation even in small-sample scenarios and has demonstrated promising results in cross-protocol and cross-attack generalisability.

\subsection{Foundation Model-based S-MAD Approaches}

Most recently, foundation models have emerged as promising tools for MAD, particularly because of their ability to generalise across
diverse attack scenarios. A first study introduced the use of Contrastive
Language-Image Pretraining (CLIP) in a zero-shot learning framework for MAD
\cite{MADCLIP}. This work examined the influence of textual prompt engineering on
detection performance and demonstrated that concise, semantically precise prompts
substantially improved the identification of morphed faces. Building on this
direction, Caldeira \textit{et al.}~\cite{caldeira2025madationfacemorphingattack} proposed the
\textit{MADation} model, which employs Low-Rank Adaptation (LoRA) to fine-tune
CLIP for enhanced task-specific discrimination. Their approach embeds LoRA layers
directly within the CLIP architecture, shifting the pre-trained feature space
towards morphing-relevant representations while simultaneously optimising a
classification head. This dual adaptation strategy improves the model’s ability to
generalise to morphs produced by previously unseen synthesis techniques.

In contrast, the proposed work introduces \textit{Residual-Statistic-Gated
Low-Rank Adapters (R-FLoRA)}, a dynamic adaptation mechanism that conditions the
transformer’s attention projections (\emph{Q}, \emph{V}) on Laplacian residual
statistics derived from each input image. Unlike prior CLIP-based frameworks,
which apply static fine-tuning or feature-space adaptation, R-FLoRA modulates the
backbone response adaptively on a per-sample basis, emphasising image-specific
high-frequency cues associated with morphing artefacts. By keeping the
foundation-scale vision transformer frozen and learning only residual-conditioned
adapters and fusion layers, the method preserves the semantic richness of CLIP
while embedding interpretable, locally aware forensic sensitivity. This design
forms a principled connection between residual analysis and foundation-model
adaptation.

Prior foundation-scale approaches adapt CLIP either by fine-tuning the full
encoder or by inserting ungated adapters that steer the global feature space. In
contrast, our proposed method uses measured Laplacian residual statistics as a conditioning
signal and injects the adaptation at the locus where attention is computed, namely
in the query and value projections. The backbone remains frozen, thereby retaining
the semantic capacity of the pre-trained model while enabling residual-aware,
sample-specific control. This combination does not appear in previous S-MAD work
and is validated empirically as a decisive factor for improved cross-dataset
performance.

While considerable progress has been achieved in S-MAD, persistent limitations in
generalisation and interpretability remain, particularly when confronting
previously unseen morphing techniques and cross-domain evaluation scenarios. To
address these challenges, this work integrates residual forensic cues with
foundation-scale vision transformer representations, achieving robust and
transferable discrimination of morphing attacks under operationally realistic
conditions.

\subsection{Novelty and Contributions}

This work presents a residual-conditioned framework for S-MAD that integrates Laplacian residual statistics with frozen foundation-scale vision transformer backbones. By explicitly coupling interpretable high-frequency cues with contextual representations learned by large pre-trained models, the proposed approach addresses generalisation and robustness challenges inherent to morphing detection. The main contributions are summarised as follows:

\begin{itemize}
    \item \textbf{Residual-conditioned backbone integration:}  
    We show that incorporating Laplacian residual statistics as an explicit conditioning signal for a frozen Vision Transformer (ViT) backbone improves generalisable morphing attack detection performance and cross-dataset generalisation, particularly for unseen morph generation techniques.

    \item \textbf{Residual-Statistic-Gated Low-Rank Adaptation module (R-FLoRA):}  
    We introduce a lightweight adaptation mechanism in which low-rank updates to the attention projections are modulated by a scalar gate derived from global residual statistics. This enables content-dependent emphasis of morph-related cues while preserving the pre-trained backbone parameters.

    \item \textbf{Residual Feature-wise Linear Modulation (Res-FiLM):}  
    A shared feature-wise modulation network is proposed to fuse patch-level residual statistics with backbone tokens, allowing local forensic evidence to be injected in a spatially aligned and parameter-efficient manner without disrupting global semantic structure.

    \item \textbf{Cross-attention pooling for adaptive aggregation:}  
    A multi-head cross-attention pooling module with learned queries is employed to aggregate fused tokens into a global representation, enabling adaptive weighting of spatially sparse and heterogeneous morphing artefacts.

    \item \textbf{Residual–Contrastive Alignment (RCA) objective:}  
    We formulate a residual-aware contrastive alignment loss that regularises the relationship between fused token representations and frozen backbone features, promoting alignment for bona fide samples while enforcing separation for morphs, thereby improving robustness under domain shift.

    \item \textbf{Comprehensive cross-dataset evaluation and ablation:}  
    The proposed framework is evaluated across four ICAO-compliant face datasets spanning multiple morph generation techniques and disjoint subject sets. Extensive ablation studies validate the individual contribution of each architectural component and loss term.


\end{itemize}

Overall, this work advances morphing attack detection by combining interpretable residual cues with foundation-model representations in a unified and parameter-efficient framework, yielding robust and generalisable performance across diverse morphing scenarios.

The remainder of this paper is organised as follows. Section~\ref{sec:Prop} presents the proposed S-MAD framework based on the R-FLoRA architecture. Section~\ref{sec:implem} details the implementation aspects, including latency analysis and computational considerations. Section~\ref{sec:DB} describes the morphing datasets used for training and evaluation. Section~\ref{sec:Exp} reports the experimental results and comparative analysis against recent state-of-the-art S-MAD methods. Section~\ref{sec:Ablation} presents the ablation study on different hyperparameters used in this work.  Section~\ref{sec:Lim} discusses the principal limitations of the study and outlines directions for future work. Finally, Section~\ref{sec:conc} summarises the main findings and concludes the paper.

\section{Proposed Method}
\label{sec:Prop}

\begin{figure*}[htp]
  \centering
  \includegraphics[width=\linewidth]{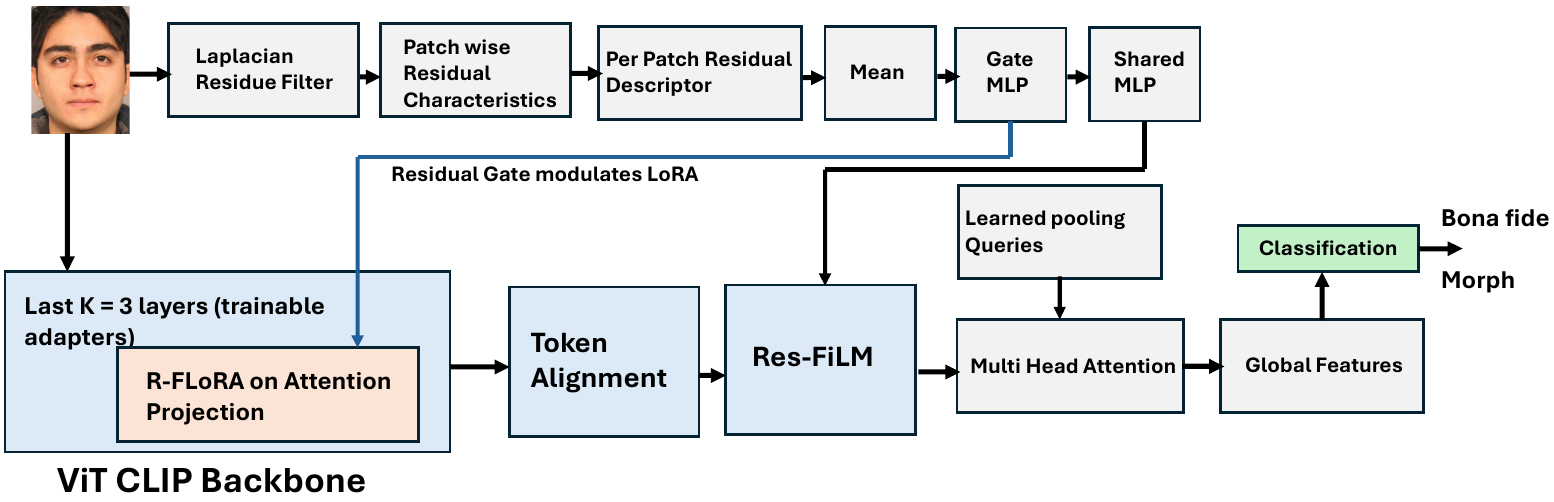}
  \caption{Overall architecture of the proposed residual-conditioned morphing attack detection framework. The input image is processed along two paths: (i) a frozen ViT backbone (CLIP~\cite{CLIP}) augmented with Residual-gated LoRA (R-FLoRA) adapters applied to the attention projections (\(Q,V\)) of the last \(k\) layers, and (ii) a Laplacian residual branch that computes patch-wise residual descriptors \(S\). The global mean of \(S\) is mapped by a gate MLP to produce a scalar residual gate that modulates the R-FLoRA updates, while the patch-level descriptors are processed by a shared MLP \(\psi\) to generate feature-wise modulation parameters \((\gamma,\beta)\) for residual feature-wise linear modulation (Res-FiLM). Backbone tokens are spatially aligned to the residual patch grid prior to Res-FiLM fusion, yielding modulated tokens \(\hat{T}\). A cross-attention pooling head with learned pooling queries \(Q_{\mathrm{pool}}\) aggregates \(\hat{T}\) into a global representation, which is finally passed to a classifier for bona fide versus morph discrimination. Training-only components are omitted for clarity.}
  \label{fig:prop}
\end{figure*}

This section introduces the proposed method for the generalised detection of single-image morphing attacks by exploiting residual information in conjunction with a frozen vision transformer backbone. An overview of the complete pipeline is shown in Fig.~\ref{fig:prop}. The framework comprises six functional components: (a) a frozen vision transformer backbone, (b) a Laplacian-based residual feature extraction branch, (c) R-FLoRA applied to the attention projections in the last $k$ transformer layers, (d) a residual feature-wise fusion module based on residual feature-wise linear modulation (Res-FiLM), (e) a cross-attention pooling head with learned pooling queries, and (f) a final classification layer. Each component is described in detail in the following subsections. Table~\ref{tab:Synbools} summarises the main symbols and notations used throughout the description of the proposed method.
\begin{table}[htp]
\centering
\caption{Summary of notation and symbols used in the proposed method and throughout the paper.}
\label{tab:Synbools}
\resizebox{\columnwidth}{!}{%
\begin{tabular}{|l|l|}
\hline
\textbf{Symbol} & \textbf{Description} \\
\hline \hline
$I$ & Input facial image \\
$R_0(x,y)$ & Raw Laplacian response at pixel $(x, y)$ \\
$R(x,y)$ & Normalised Laplacian residual, $R(x,y)=\mathrm{Norm}(R_0(x,y))$ \\
$\Omega_b$ & Local image patch or region \\
$p$ & Patch size (in pixels) \\
$N_t$ & Number of backbone tokens \\
$N_r$ & Number of residual patches \\
$M$ & Number of pixels per patch, $M=p^2$ \\
$s_i$ & Patch-wise residual summary vector $s_i=[\mu_i, \sigma_i^2, e_i]^\top$ \\
$\mu_i, \sigma_i^2, e_i$ & Mean, variance, and energy of residuals in patch $i$ \\
$S$ & Matrix of residual descriptors, $S\in\mathbb{R}^{N_r\times d_r}$ \\
$\bar{s}$ & Global mean of residual descriptors \\
$g$ & Scalar residual gate derived from $\bar{s}$ \\
$T$ & Backbone token embeddings \\
$\hat{T}$ & Residual-modulated token embeddings after Res-FiLM \\
$A, B$ & Trainable low-rank projection matrices in R-FLoRA adapters \\
$\theta_{\text{adapter}}$ & Set of all adapter parameters \\
$k$ & Number of transformer layers with active adapters \\
$Q_{\mathrm{pool}}$ & Learned pooling queries for cross-attention aggregation \\
$\mathcal{L}_{\text{RCA}}$ & Residual contrastive alignment loss \\
$\mathcal{L}_{\text{CE}}$ & Cross-entropy classification loss \\
$\mathcal{L}_{\text{total}}$ & Overall training loss, $\mathcal{L}_{\text{RCA}} + \mathcal{L}_{\text{CE}}$ \\
$y$ & Ground-truth label, $y \in \{0\ \text{(bona fide)}, 1\ \text{(morph)}\}$ \\
$\text{D-EER}$ & Detection Equal Error Rate \\
$\text{MACER}$ & Morphing Attack Classification Error Rate \\
$\text{BSCER}$ & Bona Fide Sample Classification Error Rate \\
\hline
\end{tabular}
}
\end{table}

Let the input image be
\(
I \in \mathbb{R}^{H \times W \times 3}
\)
with corresponding class label
\(
y \in \{0,1\},
\)
where $y=0$ denotes a bona fide image and $y=1$ denotes a morphed image. Morphing attack detection is formulated as a binary hypothesis testing problem:
\begin{equation}
H_0: I \sim \mathcal{D}_{\text{bona}}
\qquad \text{vs.} \qquad
H_1: I \sim \mathcal{D}_{\text{morph}},
\label{eq:hypotheses}
\end{equation}
where $\mathcal{D}_{\text{bona}}$ and $\mathcal{D}_{\text{morph}}$ denote the distributions of bona fide and morphed images, respectively. A discriminant function
\(
f_{\theta}(I) \in [0,1]
\)
is learned to estimate the posterior probability $\Pr(H_1 \mid I)$. The final decision is obtained by thresholding $f_{\theta}(I)$ according to a chosen operating point.

\subsection{Backbone (Frozen ViT)}
\label{subsec:backbone}
The proposed method is backbone agnostic and can operate with any Vision Transformer (ViT) encoder that produces a sequence of patch level token embeddings. In this work, the backbone is realised using {CLIP ViT-L/14}~\cite{CLIP}. The same framework is directly applicable to alternative foundation models such as {DINOv2 ViT-L/14}~\cite{oquab2023dinov2} and {DINOv3 ViT-L/14}~\cite{simeoni2025dinov3}, without any architectural modification.
All backbone parameters are kept frozen during training. Only the lightweight adaptation and fusion modules introduced in the subsequent subsections are optimised. These large scale ViT backbones, pre trained on diverse image text or self supervised corpora, provide rich structural and semantic representations that remain stable under limited task specific supervision. As a result, they offer a reliable feature basis for detecting the subtle spatial and textural inconsistencies introduced by morphing operations.

\paragraph{Pre-processing and tokenisation}
Let \(I \in \mathbb{R}^{H \times W \times 3}\) denote the input image. A backbone specific pre-processing operator \(\Pi_b\) resizes and normalises the image to a square resolution \(S_b \times S_b\):
\[
\tilde{I} = \Pi_b(I) \in \mathbb{R}^{S_b \times S_b \times 3},
\]
where \(\Pi_b\) applies the standard normalisation statistics associated with the selected backbone, such as CLIP statistics for CLIP or ImageNet statistics for DINO based models. The pre-processed image is partitioned into non overlapping patches of size \(p \times p\), with \(p=14\). The total number of patches is given by
\[
N = (S_b/p)^2,
\]
forming a regular patch grid
\[
g_b = (H_p, W_p) = (S_b/p,\, S_b/p).
\]
Each patch is linearly projected into an embedding space of dimension \(D\) and combined with positional encodings to produce the initial patch token matrix
\[
T_b^{(0)} \in \mathbb{R}^{N \times D}.
\]
If the backbone introduces special tokens, such as class or register tokens, these are discarded so that only spatial patch tokens are retained for subsequent processing.

\paragraph{Frozen encoder}
The transformer encoder \(F_b\), with depth \(L\), processes the patch token sequence to yield contextualised representations
\[
T = F_b^{\theta_b}\!\left(T_b^{(0)}\right) \in \mathbb{R}^{N \times D},
\]
where \(\theta_b\) denotes the frozen backbone parameters. During training, gradients do not update \(\theta_b\); instead, optimisation is restricted to the lightweight adaptation and fusion modules introduced in the following subsections. Unless stated otherwise, a unified input resolution \(S_b = S\) is adopted across all backbones to ensure a consistent backbone token grid prior to residual alignment and fusion.

\subsection{Laplacian residual features}
\label{subsec:laplacian}

Morphing operations introduce subtle but systematic high-frequency artefacts, particularly along blending boundaries and within locally edited regions. Such artefacts may persist after global appearance normalisation and are not always explicitly represented in the features extracted by a frozen backbone. To capture these cues, we extract a lightweight Laplacian-based residual representation that emphasises second-order structural and textural irregularities. The resulting residual descriptors are subsequently coupled to the frozen backbone through residual-gated low-rank adaptation and a residual feature-wise fusion mechanism.

\paragraph{Residual map}
Let \(I \in \mathbb{R}^{H \times W \times 3}\) denote an input image, and let \(\tilde I=\Pi_b(I)\in\mathbb{R}^{S_b\times S_b\times 3}\) be its backbone-specific pre-processed representation (Section~\ref{subsec:backbone}). The image domain is defined as
\(
\Omega_b=\{1,\dots,S_b\}\times\{1,\dots,S_b\}.
\)
The image is first converted to greyscale,
\begin{equation}
Y(x,y) = \mathrm{Gray}\bigl(\tilde I\bigr)(x,y), \qquad (x,y)\in\Omega_b,
\label{eq:gray}
\end{equation}
followed by Gaussian smoothing with standard deviation \(\sigma_g>0\),
\begin{equation}
Y_{\sigma_g}(x,y) = \bigl(G_{\sigma_g} * Y\bigr)(x,y),
\label{eq:gauss}
\end{equation}
after which the magnitude of the Laplacian response is computed as
\begin{equation}
R_{0}(x,y) = \bigl|\,\nabla^{2} Y_{\sigma_g}(x,y)\,\bigr|, \qquad (x,y)\in\Omega_b.
\label{eq:R0_def}
\end{equation}
A min–max normalisation yields the residual map
\begin{equation}
R(x,y) =
\frac{R_{0}(x,y) - \min\limits_{(u,v)\in\Omega_b} R_{0}(u,v)}
     {\max\limits_{(u,v)\in\Omega_b} R_{0}(u,v) - \min\limits_{(u,v)\in\Omega_b} R_{0}(u,v)}
\in [0,1].
\label{eq:R_def}
\end{equation}
The residual map emphasises edges, micro-texture, and local structural inconsistencies associated with morphing, while remaining independent of the specific backbone architecture.

\paragraph{Patch-wise residual descriptors}
Let the backbone employ a patch size \(p\) on \(\tilde I\), inducing a patch grid
\begin{equation}
g_b=(H_p,W_p) = (S_b/p,\, S_b/p),
\label{eq:grid}
\end{equation}
with a total of
\begin{equation}
N = H_p W_p = (S_b/p)^2
\label{eq:num_patches}
\end{equation}
patches. The residual map \(R\) is partitioned according to this grid, with each patch containing \(M=p^2\) residual values \(\{r_{i,m}\}_{m=1}^{M}\). For each patch \(i\in\{1,\dots,N\}\), the mean, variance, and energy are computed as
\begin{equation}
\mu_{i} = \frac{1}{M}\sum_{m=1}^{M} r_{i,m},
\label{eq:mu}
\end{equation}
\begin{equation}
\sigma^{2}_{i} = \frac{1}{M}\sum_{m=1}^{M}\bigl(r_{i,m}-\mu_{i}\bigr)^{2},
\label{eq:s2}
\end{equation}
\begin{equation}
e_{i} = \frac{1}{M}\sum_{m=1}^{M} r_{i,m}^{2}.
\label{eq:energy}
\end{equation}
The resulting per-patch descriptor is
\(s_{i}=[\mu_{i},\,\sigma^{2}_{i},\,e_{i}]^{\top}\in\mathbb{R}^{3}\),
and stacking over all patches yields
\(S=[s_{1},\dots,s_{N}]^{\top}\in\mathbb{R}^{N\times 3}\).

\paragraph{Global descriptor and downstream roles}
An image-level summary of the residual statistics is obtained by averaging the patch-wise descriptors,
\begin{equation}
\bar s = \frac{1}{N}\sum_{i=1}^{N} s_{i} \in \mathbb{R}^{3}.
\label{eq:sbar}
\end{equation}
The global descriptor \(\bar s\) is passed through a small multi-layer perceptron \(\phi:\mathbb{R}^{3}\rightarrow\mathbb{R}\), followed by a sigmoid non-linearity to produce a scalar residual gate \(g=\mathrm{sigm}(\phi(\bar s))\). This gate modulates the low-rank adaptation modules applied to the attention projections in the final \(k\) backbone layers (Section~\ref{subsec:rflora}). In parallel, the patch-wise residual descriptors \(S\) are spatially aligned to the backbone token grid and provide the conditioning signal for the residual feature-wise linear modulation (Res-FiLM) fusion stage and subsequent classification.

It is worth noting that the Laplacian operator acts as a simple high-pass filter, responding strongly to rapid intensity transitions and subtle structural inconsistencies introduced by blending or retouching. The statistics in~\eqref{eq:mu}–\eqref{eq:energy} capture the strength and variability of these cues at the patch level, providing a stable and interpretable residual representation that complements the contextual token embeddings produced by the frozen backbone, without requiring supervision beyond image-level labels.

\subsection{R-FLoRA adapters}
\label{subsec:rflora}

Building on the residual features defined in Section~\ref{subsec:laplacian}, we introduce R-FLoRA to modulate a frozen Vision Transformer using a small number of trainable parameters. The core idea is to inject low-rank updates into the attention projections and to scale these updates using a gate derived from global residual statistics. This design enables the backbone to adapt to morphing specific cues without modifying its pre-trained weights, while remaining compatible with CLIP ViT-L/14~\cite{CLIP}, DINOv2 ViT-L/14~\cite{oquab2023dinov2}, and DINOv3 ViT-L/14~\cite{simeoni2025dinov3} backbones.

Adaptation is restricted to the attention projections \(Q\) and \(V\), as these components directly control the spatial focus and content aggregation of the model. Modifying output projections or feed-forward blocks increases parameter count without providing comparable control over attention behaviour and may destabilise optimisation when the encoder is frozen. The adapters are applied only in the final \(k=3\) transformer layers, targeting high-level, task-specific interactions while preserving earlier, general-purpose representations.

\paragraph{Residual-gated low-rank updates}
Let \(T \in \mathbb{R}^{N \times D}\) denote the patch token sequence produced by the frozen backbone (Section~\ref{subsec:backbone}), and let the transformer encoder have depth \(L\). For a given transformer layer \(\ell\), the standard attention projections are:
\begin{equation}
q_{\ell} = T_{\ell}\, W^{(q)}_{\ell}, 
\qquad
v_{\ell} = T_{\ell}\, W^{(v)}_{\ell},
\label{eq:std_proj}
\end{equation}
where \(W^{(q)}_{\ell}, W^{(v)}_{\ell} \in \mathbb{R}^{D \times D}\) are frozen weight matrices. From the Laplacian residual branch, we obtain the patch-wise residual descriptors \(s_i \in \mathbb{R}^{3}\) and their image-level mean:
\begin{equation}
\bar{s} = \frac{1}{N}\sum_{i=1}^{N} s_i \in \mathbb{R}^{3}.
\label{eq:sbar_rflora}
\end{equation}
A lightweight gating network \(\phi:\mathbb{R}^{3}\rightarrow\mathbb{R}\) maps \(\bar{s}\) to a scalar gate shared across all tokens and attention heads for the image,
\begin{equation}
g = \mathrm{sigm}\!\bigl(\phi(\bar{s})\bigr) \in (0,1),
\label{eq:gate}
\end{equation}
where \(\mathrm{sigm}(\cdot)\) denotes the logistic activation.

For each projection \(P \in \{q,v\}\) at layer \(\ell\), R-FLoRA introduces trainable low-rank factors:
\begin{equation}
A^{(P)}_{\ell} \in \mathbb{R}^{D \times r},
\qquad
B^{(P)}_{\ell} \in \mathbb{R}^{r \times D},
\label{eq:AB}
\end{equation}
with rank \(r \ll D\) and a fixed scaling factor \(\alpha>0\). The resulting gated projection is given by:
\begin{equation}
T_{\ell}\,\widetilde{W}^{(P)}_{\ell}
=
T_{\ell}\, W^{(P)}_{\ell}
+
\alpha\, g \, T_{\ell} A^{(P)}_{\ell} B^{(P)}_{\ell},
\qquad
\ell \in \{L-k+1,\dots,L\}.
\label{eq:rflora_update}
\end{equation}
Only the low-rank factors \(A^{(P)}_{\ell}\), \(B^{(P)}_{\ell}\), and the parameters of the gating network \(\phi\) are updated during training; all backbone weights remain frozen. The adapters are applied to both \(Q\) and \(V\) projections in the final \(k\) layers. For backbones that implement a fused \(QKV\) projection, the adapters are attached to the corresponding \(Q\) and \(V\) parameter slices.

\paragraph{Complexity and design rationale}
R-FLoRA operates directly on patch tokens after special tokens are removed and is independent of the specific backbone architecture. The parameter increase per layer and per projection is \(2Dr\), with a comparable number of multiply–add operations in the low-rank path. Restricting the adapters to the final \(k\) layers keeps the computational and memory overhead modest relative to the frozen encoder, while the image-conditioned gate \(g\) provides controlled, content-dependent modulation that is important for capturing morphing related artefacts.

\subsection{Fusion adapter (Res-FiLM)}
\label{subsec:fusion}

The backbone tokens \(T \in \mathbb{R}^{N_t \times D}\) encode global semantic context, whereas morphing artefacts are predominantly local and high-frequency. We therefore condition the backbone representations using lightweight local signals derived from the Laplacian residual statistics \(S \in \mathbb{R}^{N_r \times 3}\) (Section~\ref{subsec:laplacian}). To this end, we adopt a feature-wise modulation mechanism that scales and shifts each token channel based on patch-level residual cues. This {Residual Feature-wise Linear Modulation (Res-FiLM)} is parameter efficient, stable under frozen backbones, and requires only a small shared multilayer perceptron, without introducing additional attention or projection layers.

Let
\(
T=[t_{1},\dots,t_{N_t}]^{\top}\in\mathbb{R}^{N_t\times D}
\)
denote the backbone token sequence after R-FLoRA modulation in the final \(k\) transformer layers (Section~\ref{subsec:rflora}). From the Laplacian branch, we obtain patch-wise residual descriptors
\(
S=[s_{1},\dots,s_{N_r}]^{\top}\in\mathbb{R}^{N_r\times 3},
\)
where each
\(s_{i}=[\mu_{i},\sigma^{2}_{i},e_{i}]^{\top}\)
is defined in~\eqref{eq:mu}–\eqref{eq:energy}. Since the residual descriptors may be computed on a patch grid that differs from the backbone token grid, the token sequence \(T\) is spatially aligned to the residual grid prior to fusion, yielding an aligned token sequence \(T' \in \mathbb{R}^{N_r \times D}\).

\paragraph{Architecture of the shared modulation network}
A small shared multilayer perceptron maps each residual descriptor \(s_i\) to feature-wise modulation parameters \(\gamma_i,\beta_i \in \mathbb{R}^{D}\):
\begin{equation}
h_i = \mathrm{SiLU}\!\bigl(W_1 s_i + b_1\bigr), 
\qquad
[\gamma_i,\beta_i] = W_2 h_i + b_2,
\label{eq:psi_mlp}
\end{equation}
where \(W_1\!\in\!\mathbb{R}^{128\times 3}\), \(b_1\!\in\!\mathbb{R}^{128}\), \(W_2\!\in\!\mathbb{R}^{2D\times 128}\), and \(b_2\!\in\!\mathbb{R}^{2D}\). The activation function \(\mathrm{SiLU}(\cdot)\) denotes the Sigmoid Linear Unit. The first \(D\) outputs correspond to \(\gamma_i\) and the remaining \(D\) outputs to \(\beta_i\). All parameters \((W_1,b_1,W_2,b_2)\) are shared across patches \(i=1,\dots,N_r\).

\paragraph{Res-FiLM modulation}
Each aligned token \(t'_i \in \mathbb{R}^{D}\) is modulated as
\begin{equation}
\hat t_{i} = \bigl(1+\tanh(\gamma_{i})\bigr)\odot t'_{i} + \beta_{i},
\qquad i=1,\dots,N_r,
\label{eq:film_update_single}
\end{equation}
where \(\odot\) denotes the Hadamard product and \(\tanh(\gamma_i)\) bounds the scaling factor for numerical stability. Stacking across patches yields the fused token matrix
\(
\hat T=[\hat t_{1},\dots,\hat t_{N_r}]^{\top}\in\mathbb{R}^{N_r\times D}.
\)
Residual token embeddings are not used in this configuration; fusion relies solely on the patch-wise residual statistics \(S\).

The fused tokens \(\hat T\) retain the global semantic information of the frozen backbone while being locally conditioned by Laplacian-driven residual cues. These representations are subsequently aggregated by the cross-attention pooling head (Section~\ref{subsec:head}) to produce the final morphing decision score.

\subsection{Cross-Attention Pooling and Classification}
\label{subsec:head}

The fused tokens \(\hat T \in \mathbb{R}^{N_r \times D}\) (Section~\ref{subsec:fusion}) encode global semantic context from the frozen backbone while being locally conditioned by Laplacian-driven residual cues. Since morphing artefacts are spatially sparse and vary in location and strength, fixed pooling strategies such as mean or max pooling treat all patches uniformly and may dilute discriminative evidence. To address this limitation, we employ a cross-attention pooling head that learns content-adaptive weights over patches, enabling the model to focus on regions exhibiting residual-conditioned inconsistencies while preserving holistic context.

Let \(\hat T=[\hat t_1,\dots,\hat t_{N_r}]^{\top}\), with \(\hat t_i\in\mathbb{R}^{D}\). The pooling head is parameterised by a small set of learned pooling queries
\(
Q_{\mathrm{pool}}=\{q_m\}_{m=1}^{M}, \; q_m\in\mathbb{R}^{D},
\)
which are trainable parameters independent of the input image. These queries probe the fused token sequence via cross-attention to extract a compact global representation.

\paragraph{Cross-attention pooling}
For each pooling query \(q_m\), key and value projections are formed from the fused tokens, and a query projection is applied to \(q_m\):
\begin{equation}
K = \hat T\, W^{(K)} \in \mathbb{R}^{N_r\times d_k}, 
\qquad
V = \hat T\, W^{(V)} \in \mathbb{R}^{N_r\times d_v},
\label{eq:kv_proj}
\end{equation}
\begin{equation}
q_m' = q_m\, W^{(Q)} \in \mathbb{R}^{d_k},
\label{eq:q_proj}
\end{equation}
where \(W^{(K)}\in\mathbb{R}^{D\times d_k}\), \(W^{(V)}\in\mathbb{R}^{D\times d_v}\), and \(W^{(Q)}\in\mathbb{R}^{D\times d_k}\) are trainable projection matrices.

With \(k_i\) and \(v_i\) denoting the \(i\)-th rows of \(K\) and \(V\), respectively, the attention weights and pooled feature for query \(q_m\) are given by
\begin{equation}
a_{m,i} =
\frac{\exp\!\bigl(\langle q_m',\,k_i\rangle / \sqrt{d_k}\bigr)}
     {\sum_{j=1}^{N_r}\exp\!\bigl(\langle q_m',\,k_j\rangle / \sqrt{d_k}\bigr)},
\qquad i=1,\dots,N_r,
\label{eq:attn_weights}
\end{equation}
\begin{equation}
\mathbf{g}_m = \sum_{i=1}^{N_r} a_{m,i}\, v_i \in \mathbb{R}^{d_v}.
\label{eq:global_feature}
\end{equation}
Each \(\mathbf{g}_m\) represents a content-adaptive aggregation of the fused tokens, emphasising regions where residual-conditioned representations indicate morph-related evidence.

\paragraph{Multi-head pooling and classification}
In practice, cross-attention pooling is implemented using a multi-head formulation with \(H=8\) attention heads and a learned query bank \(Q_{\mathrm{pool}}\). Each head computes pooled features as in~\eqref{eq:attn_weights}–\eqref{eq:global_feature}, yielding head-specific outputs that are concatenated and linearly projected to form a global feature vector \(\mathbf{g}\). A linear classifier then maps \(\mathbf{g}\) to a logit,
\begin{equation}
z = w^{\top} \mathbf{g} + b,
\qquad
s = \mathrm{sigm}(z) \in [0,1],
\label{eq:classifier}
\end{equation}
where \(w\in\mathbb{R}^{d_v}\) and \(b\in\mathbb{R}\) are trainable parameters. The scalar score \(s\) is used for morphing attack detection at inference. No class token is introduced, and no additional pooling is applied prior to cross-attention.

\subsection{Training objective}
\label{subsec:loss}

Morphing artefacts are sparse, heterogeneous, and spatially localised. While an image-level classification loss is effective for global discrimination, it may underutilise fine-grained patch-level evidence. We therefore combine standard objectives, including image-level binary cross-entropy (BCE) with label smoothing, sparse Multiple Instance Learning (MIL), and a Total Variation (TV) spatial prior, with a residual-aware contrastive alignment term referred to as {Residual–Contrastive Alignment (RCA)}. The RCA term encourages residual-conditioned fused token representations to remain close to the frozen backbone representations for bona fide images, while enforcing a separation margin for morphs. Together, these losses balance global decision confidence with localised evidence, promote spatial coherence, and stabilise residual-driven adaptation across diverse morphing styles.

\paragraph{Image-level classification}
Let \(z\in\mathbb{R}\) denote the logit produced by the cross-attention pooling head (Section~\ref{subsec:head}), and let \(y\in\{0,1\}\) be the corresponding ground-truth label. With label-smoothing coefficient \(\varepsilon\in[0,1)\), the primary classification loss is defined as
\begin{equation}
\begin{aligned}
\tilde y &= (1-\varepsilon)\,y+\tfrac{\varepsilon}{2},\\
\mathcal{L}_{\mathrm{main}}
&= -\,\tilde y\log\mathrm{sigm}(z)
-(1-\tilde y)\log\!\bigl(1-\mathrm{sigm}(z)\bigr).
\end{aligned}
\label{eq:loss_main}
\end{equation}

\paragraph{Sparse Multiple Instance Learning (MIL) (training only)}
Let \(\ell\in\mathbb{R}^{N_r}\) denote patch-level logits predicted by the auxiliary MIL head defined on the residual patch grid. With \(k=\max(1,\lfloor\rho N_r\rfloor)\), the positive and negative instance scores are computed as
\begin{equation}
\begin{aligned}
s_{\text{pos}} &= \frac{1}{k}\sum_{i\in \mathrm{TopK}(\ell,k)} \ell_i,\\
s_{\text{neg}} &= \frac{1}{k}\sum_{i\in \mathrm{TopK}(-\ell,k)} (-\ell_i),\\
z_{\mathrm{mil}} &= y\, s_{\text{pos}} + (1-y)\, s_{\text{neg}}.
\end{aligned}
\label{eq:mil_pool}
\end{equation}
The corresponding loss is
\begin{equation}
\mathcal{L}_{\mathrm{mil}}
=-\,\tilde y\log\mathrm{sigm}(z_{\mathrm{mil}})
-(1-\tilde y)\log\!\bigl(1-\mathrm{sigm}(z_{\mathrm{mil}})\bigr),
\label{eq:loss_mil}
\end{equation}
where the MIL head is used exclusively during training and removed at inference.

\paragraph{Residual–Contrastive Alignment (RCA)}
Let \(T'\) and \(\hat T\in\mathbb{R}^{N_r\times D}\) denote the aligned backbone tokens and the residual-modulated tokens on the residual grid, respectively (Section~\ref{subsec:fusion}). For each patch \(i\), we define the cosine distance
\begin{equation}
d_i = 1-
\Bigl\langle 
\tfrac{T'_i}{\|T'_i\|_2},\,
\tfrac{\hat T_i}{\|\hat T_i\|_2}
\Bigr\rangle 
\in [0,2],
\qquad i=1,\dots,N_r.
\label{eq:rca_dist}
\end{equation}
With a batch-adaptive margin \(m\), computed as a fixed quantile over morph samples, and a warm-up factor \(w\in[0,1]\), the RCA loss is defined as
\begin{equation}
\mathcal{L}_{\mathrm{rca}}
=\frac{1}{N_r}\sum_{i=1}^{N_r}
\Bigl[(1-y)\,d_i + y\,\phi(m-d_i)\Bigr],
\qquad
\mathcal{L}_{\mathrm{rca,w}}=w\,\mathcal{L}_{\mathrm{rca}},
\label{eq:loss_rca}
\end{equation}
where \(\phi(\cdot)\) denotes the Softplus function. Intuitively, RCA aligns bona fide token embeddings with their frozen-backbone counterparts while encouraging separation for morph tokens.

\paragraph{Total Variation (TV) prior}
Morphing artefacts typically manifest as spatially coherent regions rather than isolated patch-level anomalies. To promote spatial smoothness in the patch-logit map while preserving sharp transitions, an anisotropic total-variation penalty \(\mathcal{L}_{\mathrm{tv}}\) is applied. This regulariser suppresses spurious isolated activations arising from noise or stochastic optimisation and biases the MIL supervision towards contiguous, morph-consistent regions.

\paragraph{Overall objective}
The complete training objective is given by
\begin{equation}
\mathcal{L}
=\mathcal{L}_{\mathrm{main}}
+\lambda_{\mathrm{mil}}\,\mathcal{L}_{\mathrm{mil}}
+\lambda_{\mathrm{tv}}\,\mathcal{L}_{\mathrm{tv}}
+\lambda_{\mathrm{rca}}\,\mathcal{L}_{\mathrm{rca,w}},
\label{eq:loss_total}
\end{equation}
where all weighting coefficients \(\lambda_{\bullet}\geq 0\) are selected based on validation performance. In all experiments, we set 
$\lambda_{\mathrm{mil}}=0.6$, 
$\lambda_{\mathrm{tv}}=0.05$, and 
$\lambda_{\mathrm{rca}}=0.05$, 
selected based on validation performance. A comprehensive sensitivity analysis is provided in the Supplementary Material.

\paragraph{Notation recap}
\(N_r\): number of residual patches;  
\(D\): token dimensionality;  
\(\mathrm{sigm}\): logistic function;  
\(\mathrm{TopK}\): operator returning the indices of the largest elements;  
\(\|\cdot\|_2\): Euclidean norm.

\section{Implementation Details and Latency}
\label{sec:implem}

The proposed S\textnormal{-}MAD model is implemented in PyTorch (v2.3), with mixed-precision training and evaluation enabled via \texttt{torch.cuda.amp}. The backbone is incorporated from \texttt{openai/clip-vit-large-patch14-336}, operating at \(336\times336\) resolution with all backbone parameters frozen during training. The trainable components are limited to lightweight adaptation, fusion, and classification modules, yielding approximately \(9.3\) million trainable parameters out of the \(\sim 427\) million total in CLIP ViT-L/14. These include: RFLoRA low-rank adapters (Q and V projections in the final \(k{=}3\) layers, \(\approx 7.6\)~M parameters), the Res-FiLM fusion MLP (\(\approx 1.4\)~M), and the cross-attention pooling head with learned queries (\(\approx 0.3\)~M). This corresponds to a \(2.1\%\) trainable fraction of the full backbone.

Each forward pass consists of: (i) residual map extraction via Laplacian filtering, (ii) feature propagation through the frozen CLIP encoder, (iii) low-rank modulation via R-FLoRA adapters, and (iv) Res-FiLM fusion followed by cross-attention pooling and classification. The total computational cost of the GPU model, measured using \texttt{fvcore}, is approximately \(182.41\)~GFLOPs per image at \(336{\times}336\) resolution. This estimate excludes the lightweight CPU-based residual pre-processing, whose cost is negligible relative to the backbone inference. The overall complexity is consistent with that of a ViT-L/14-scale transformer and confirms the efficiency of the additional forensic components.

\paragraph*{Training setup}
Training is performed for \(40\) epochs using the Adam optimiser with an initial learning rate of \(2\times10^{-4}\), weight decay \(0.05\), and gradient accumulation over four steps. The batch size is \(8\), and an exponential moving average (EMA, \(\tau{=}0.999\)) is applied for parameter stabilisation. A curriculum schedule progressively increases residual noise and JPEG degradation probabilities across epochs. The residual path employs Laplacian-magnitude filtering with Gaussian pre-smoothing \((\sigma\in[0.6,1.6])\). Data augmentation includes horizontal flips, mild colour jitter, Gaussian blur, and resampling distortions, which simulate camera and compression artefacts without altering intrinsic forensic cues.

\paragraph*{Evaluation and latency}
Inference latency is benchmarked on an NVIDIA L40 (48~GB) GPU using CUDA event timers with 30 warm-up and 300 timed runs, at batch size \(1\) and \(336{\times}336\) input resolution. The mean per-image latency is \(12.99\)~ms, comprising residual pre-processing (\(0.10\)~ms) and GPU model inference (\(12.91\)~ms), corresponding to a throughput of approximately \(77.0\) images per second. The residual computation and adapter-based fusion together account for less than \(1\%\) of the total latency, confirming that the added forensic components impose negligible overhead relative to the frozen backbone.

\paragraph*{Resource efficiency}
By keeping the ViT encoder frozen and introducing conditional low-rank adaptation, S\textnormal{-}MAD achieves strong discriminative performance with minimal computational and memory overhead. The model supports half-precision inference (\texttt{torch.float16}) without observable accuracy degradation and is amenable to further optimisation via quantisation or distillation. The Laplacian residual pipeline can operate entirely on CPU if required, enabling hybrid CPU–GPU execution with an end-to-end latency of approximately \(13\)~ms per image on modern hardware, supporting real-time deployment on standard GPU accelerators.


\begin{figure*}[htp]
  \centering
  \includegraphics[width=\linewidth]{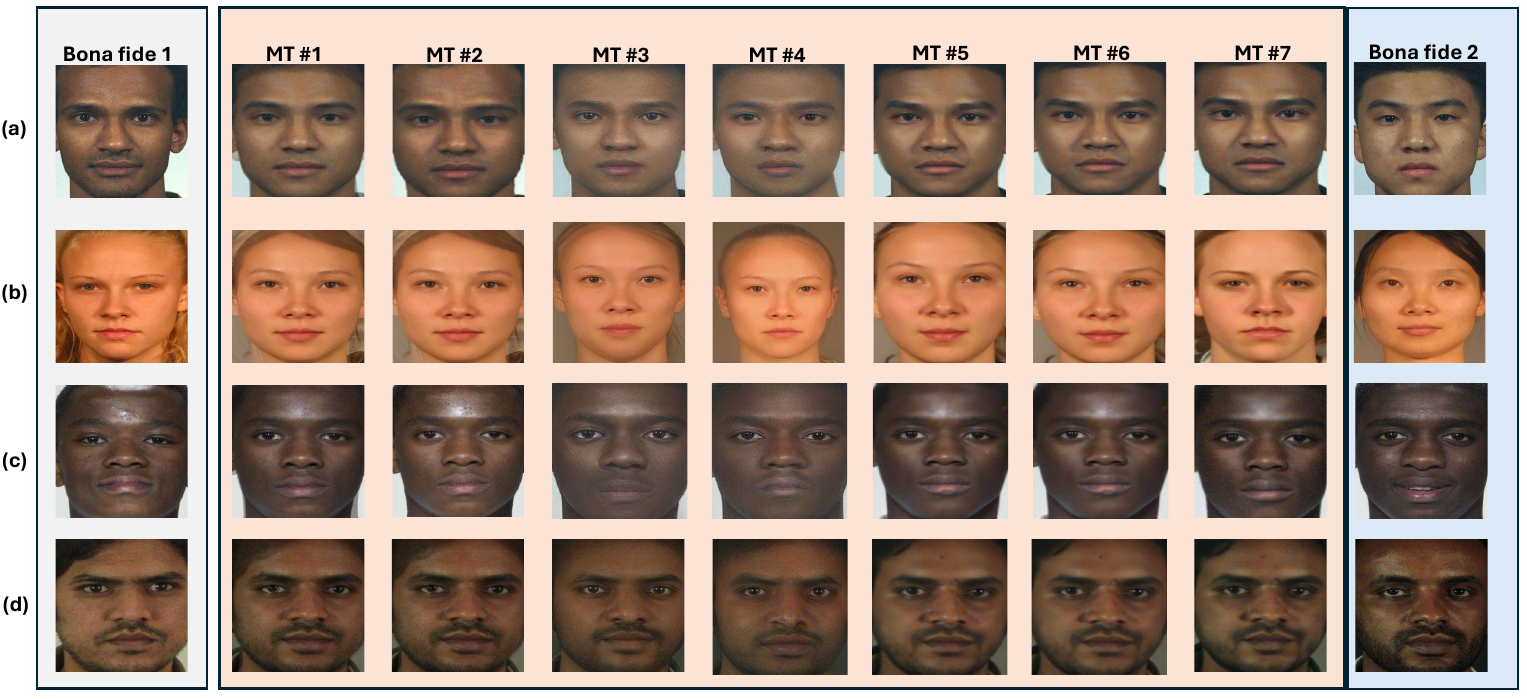}
  \caption{Example facial images from the four different face morphing datasets with seven different Morphing Types (MT) (a) MD\#1 (b) MD\#2 (c) MD\#3 (d) MD\#4}
  \label{fig:db}
\end{figure*}
\section{Morphing Datasets}
\label{sec:DB}
This section provides detailed information on the four morphing datasets used to benchmark the performance of the proposed and existing S-MAD methods. The morphing datasets were derived from four publicly available face databases: FERET~\cite{phillips1998feret}, FRGC~\cite{FRGC_DB}, FRILL~\cite{FRILL_DB}, and MS40~\cite{MS40_DB}. All bona fide images from these datasets were first processed to ensure compliance with ICAO standards~\cite{ICAO-9303-p1-2015}. Morphing was subsequently carried out following the guidelines in~\cite{Raghavendra-FaceMorphingVersusFaceAveraging-IJCB-2017}, ensuring that morph pairs were formed between subjects of the same gender and ethnicity. Candidate subjects for morphing were selected based on their similarity scores obtained using the MagFace face recognition system~\cite{magface}.

\begin{table}[htp]
\centering
\caption{Abbreviations of different morphing generation techniques used in this work.}
\label{tab:MADMEthods}
\resizebox{1.0\textwidth}{!}{
\begin{tabular}{|l|l|}
\hline
\rowcolor[HTML]{EFEFEF} 
\textbf{Morphing Generation Methods}                                                  & \textbf{Abbreviation} \\ \hline \hline
Landmark based morphing generation \cite{Raghavendra-FaceMorphingVersusFaceAveraging-IJCB-2017}                            & MT \#1                \\ \hline
Landmark based morphing generation with post-processing \cite{Magicmorph-morphing-software}       & MT \#2                \\ \hline
MIPGAN-1 based morphing generation \cite{zhang-MIPGAN-TBIOM-2021}                            & MT \#3                \\ \hline
MIPGAN-2 based morphing generation \cite{zhang-MIPGAN-TBIOM-2021}                            & MT \#4                \\ \hline
Diffusion Model based morphing generation  (MorDiff)\cite{MoDiff}                     & MT \#5                \\ \hline
Diffusion Model with identity loss based morphing generation (PIPE) \cite{PIPE} & MT \#6                \\ \hline
Diffusion Model with greedy based morphing generation (Greedy) \cite{blasingame_greedy_dim} & MT \#7                \\ \hline
\end{tabular}
}
\end{table}

In this work, seven distinct morph generation techniques were employed, comprising both traditional landmark-based methods and generative models. The handcrafted techniques included landmark-based morphs with~\cite{Raghavendra-FaceMorphingVersusFaceAveraging-IJCB-2017} and without post-processing~\cite{Magicmorph-morphing-software}. The generative approaches included adversarial and diffusion-based models~\cite{zhang-MIPGAN-TBIOM-2021, MoDiff, PIPE, blasingame_greedy_dim}. Table~\ref{tab:MADMEthods} provides a summary of the different Morphing Techniques MT\#1–MT\#7 and their corresponding generation methods. 

The handcrafted methods directly combine pixel information from bona fide images, thereby preserving identity, shape, and texture fidelity. As a result, they often produce morphs with strong visual and biometric similarity to the contributing subjects. In contrast, generative models optimise image synthesis within a latent space, which can lead to slight reductions in texture realism compared with handcrafted approaches. The statistics and composition of the four morphing datasets are described below.

\subsection*{Morphing Dataset (MD) \#1 (FERET)}
This dataset was created from the FERET database~\cite{phillips1998feret}. It includes 766 unique subjects partitioned by gender and ethnicity. Morphs were generated using all seven morph generation techniques described above. The dataset contains 2,003 bona fide images and 13,753 morph images. Representative examples are shown in Figure~\ref{fig:db}~(a).

\subsection*{Morphing Dataset (MD) \#2 (FRGC)}
This dataset was constructed using the FRGC database~\cite{FRGC_DB}. A total of 150 unique subjects were selected and partitioned by gender and ethnicity, followed by morph generation using all seven techniques. It comprises 1,276 bona fide and 6,365 morph images. Example illustrations appear in Figure~\ref{fig:db}~(b).

\subsection*{Morphing Dataset (MD) \#3 (FRLL)}
The FRLL database~\cite{FRILL_DB} was used to build this dataset. Although FRLL includes a small subset of existing morphs, additional morphs were produced using all seven techniques to maintain consistency with the other datasets. The dataset includes 102 bona fide and 7,784 morph images. Figure~\ref{fig:db}~(c) provides sample images.

\subsection*{Morphing Dataset (MD) \#4 (MS40)}
This dataset was derived from the MS40 database~\cite{MS40_DB}. Only visible-spectrum images acquired under controlled studio conditions were selected to simulate a passport-style application requiring high-quality facial imagery. The dataset contains 498 bona fide and 2,660 morph images. Representative examples are shown in Figure~\ref{fig:db}~(d).

\noindent
The four morphing datasets used in this study were constructed exclusively from \textit{real facial images} rather than synthetically generated faces. This design ensures that the datasets capture genuine biometric variability and realistic imaging characteristics such as illumination, texture, and sensor noise, which are important for evaluating morphing detection systems intended for operational deployment. Synthetic or GAN-based faces, though visually plausible, tend to exhibit artefacts or non-biometric regularities that could bias the evaluation of generalisable morphing detection performance.

All bona fide images selected were ICAO-compliant to reflect the quality specifications required for official identity enrolment. The ICAO standard defines strict criteria for pose, illumination, and background uniformity, ensuring passport-quality images suitable for machine-readable identity verification. This decision aligns the datasets with real-world identity management systems, where ICAO-conformant images are standard for passports, national identity cards, and electronic visa documents. Many countries now accept only \textit{digital} facial photographs for such applications, emphasising consistent quality and secure electronic submission over print–scan workflows.

Accordingly, only the digital versions of images were included in our datasets. Producing high-quality print–scan counterparts at this scale would require dedicated imaging hardware, repeated calibration, and manual inspection, making the process both time-consuming and costly. Since the current study focuses on detecting morphing artefacts within the image domain rather than analysing sensor-induced distortions, the use of digital ICAO-compliant images provides a realistic and efficient basis for large-scale benchmarking of S-MAD methods.

\section{Experiments and Results}
\label{sec:Exp}
This section details the experimental analysis of the proposed approach in comparison with established Single image based morphing attack detection (S-MAD) techniques. Detection performance was assessed according to the ISO/IEC SC 37 30107 standards \cite{ISO-IEC-30107-3-PAD-metrics-170227}, ensuring adherence to internationally recognised evaluation criteria.

The principal metrics used in these experiments are the Morphing Attack Classification Error Rate (MACER) and the Bona Fide Sample Classification Error Rate (BSCER). MACER quantifies the fraction of morphing attack samples incorrectly accepted as genuine, while BSCER measures the proportion of bona fide presentation samples mistakenly classified as morphing attacks. For a comprehensive assessment, the Detection Equal Error Rate (D-EER) is also computed. D-EER represents the operating point at which MACER and BSCER are equal, offering a balanced perspective on the overall detection capability of each method.

\subsection{Performance Evaluation Protocol}
In this study, an unseen dataset evaluation protocol is employed to assess the generalisation capability of both the proposed and established S‑MAD methods. Of the four morphing datasets utilised, MD\#1 and MD\#2 contain sufficiently large sets of bona fide and morph images; these serve as sources for training and validation. MD\#3 and MD\#4 are reserved exclusively as evaluation sets, ensuring that test images remain entirely separate from those used during method development.

Two distinct experiments have been designed. In the first experiment (\textbf{Experiment \#1}), training is performed on MD\#1, and validation utilises MD\#2. The validation set guides the selection of hyperparameters and determination of operating thresholds. Final detection performance is then measured independently on MD\#3 and MD\#4. In the second experiment (\textbf{Experiment \#2}), the training set is switched to MD\#2, with validation on MD\#1. As before, hyperparameter selection and threshold setting are based on the validation results, and evaluation is conducted strictly on MD\#3 and MD\#4.

This protocol ensures that the assessment of S‑MAD algorithms reflects their effectiveness on completely unseen data, providing a robust test of generalisability for morphing attack detection under realistic and challenging conditions.
\subsection{Rationale for the evaluation protocol}
The chosen train–validation protocol is designed to explicitly assess cross-dataset generalisation rather than in-domain optimisation. MD\#2 is used for training due to its large scale, controlled acquisition conditions, and wide subject coverage, which support stable learning of the proposed residual-guided adaptation without overfitting. MD\#1 is employed exclusively for validation and cross-dataset evaluation because it exhibits a substantial domain shift in terms of sensors, acquisition periods, and demographic composition. This configuration reflects a realistic deployment scenario in which a morphing attack detector trained on one operational dataset must generalise to previously unseen data. While alternative dataset permutations are possible, the selected protocol follows established S-MAD evaluation practice and provides a challenging yet reproducible benchmark for assessing robustness.

\subsection{Results and Discussion}
This section presents the quantitative evaluation of both the proposed and existing S-MAD methods. Detection performance is reported as the average across the various morph types in the evaluation datasets. To support robust interpretation, 95\% confidence intervals are provided, reflecting the statistical significance and reliability of the results.

The selection of existing methods was guided by recent progress in S-MAD techniques. Specifically, the chosen nine approaches represent state-of-the-art strategies encompassing a diverse range of machine learning paradigms. These include methods based on diffusion probabilistic models~\cite{ivanovska2023mad_ddpm}, self-supervised  architecture~\cite{ivanovska2025selfmad}, Localised Features \cite{qin2022face}, Camera Noise \cite{zhang2018face}, teacher–student distillation architecture~\cite{RiaCVIP-2025}, few-shot learning framework~\cite{tapia2025single}, foundation models and transfer learning~\cite{caldeira2025madationfacemorphingattack}, and transformer-based~\cite{zhang2024ViT}. Each of these techniques has been selected to ensure coverage of the most prominent machine learning advances shaping recent S-MAD research, and to provide a rigorous and representative benchmark for the evaluation of the proposed approach.

\subsubsection{\textbf{Quantitative Results: train: MD\#1  and Validation: MD\#2.}}
\begin{table*}
\caption{Quantitative results of S-MAD techniques when evaluated on MD\#3 database (Train:Val MD\#1:MD\#2). BSCER values are averaged over Morph Types (MTs) (mean ±95\% CI).}
\label{tab:frill_feret_frgc_main}
\begin{tabular}{lcccc}
\toprule
Method & D-EER\% & BSCER@MACER=1\% & BSCER@MACER=5\% & BSCER@MACER=10\% \\
\midrule
\textbf{Proposed Method} & \textbf{7.14} $\pm$ \textbf{0.45} & \textbf{18.21} $\pm$ \textbf{4.49} & \textbf{7.70} $\pm$ \textbf{3.51} & \textbf{3.78} $\pm$ \textbf{1.96} \\
Diffusion probabilistic models~\cite{ivanovska2023mad_ddpm} & 57.07 $\pm$ 1.65 & 100.00 $\pm$ 0 & 99.58 $\pm$ 0.28 & 98.04 $\pm$ 0.45 \\
Self-Supervision~\cite{ivanovska2025selfmad} & 30.89 $\pm$ 1.76 & 65.27 $\pm$ 21.99 & 51.26 $\pm$ 25.07 & 46.78 $\pm$ 24.93 \\
Distil-MAD~\cite{RiaCVIP-2025} & 16.57 $\pm$ 0.81 & 60.78 $\pm$ 2.67 & 38.24 $\pm$ 1.56 & 28.43 $\pm$ 3.42 \\
Few-Shot MAD (MobileNet)~\cite{tapia2025single} & 33.60 $\pm$ 1.82 & 80.39 $\pm$ 9.66 & 69.89 $\pm$ 12.75 & 59.38 $\pm$ 15.69 \\
Few-Shot MAD (ResNet)~\cite{tapia2025single} & 48.85 $\pm$ 2.31 & 94.96 $\pm$ 0.56 & 92.30 $\pm$ 1.26 & 86.97 $\pm$ 3.64 \\
MADation \cite{caldeira2025madationfacemorphingattack} & 49.60 $\pm$ 1.53 & 65.27 $\pm$ 26.89 & 56.30 $\pm$ 26.89 & 47.34 $\pm$ 17.93 \\
ViTSVM~\cite{zhang2024ViT} & 14.95 $\pm$ 1.36 & 44.40 $\pm$ 19.48 & 28.43 $\pm$ 17.23 & 21.29 $\pm$ 14.29 \\
LocalMorph~\cite{qin2022face} & 12.74 $\pm$ 1.34 & 22.13 $\pm$ 8.48 & 10.78 $\pm$ 3.33 & 6.86 $\pm$ 2.47 \\
FS-SPN~\cite{zhang2018face} & 29.41 $\pm$ 1.45 & 89.21 $\pm$ 8.54 & 72.54 $\pm$ 2.45 & 51.96 $\pm$ 5.47 \\
\bottomrule
\end{tabular}
\end{table*}
Table~\ref{tab:frill_feret_frgc_main} summarises the quantitative performance of the proposed and existing S-MAD methods on the FRILL dataset, including 95\% confidence intervals for each reported metric. 
The quantitative results in Table~\ref{tab:frill_feret_frgc_main} demonstrate that the proposed S-MAD method substantially outperforms all comparators on the MD\#3 dataset in a cross-dataset setting (Train: MD\#1, Validation: MD\#2, Test: MD\#3). The proposed method achieves a D-EER of $7.14 \pm 0.45$, and consistently strong BSCER values at all operating points, with statistically significant improvement over both classical and state-of-the-art approaches.
The superior performance of the proposed method is attributed to several key architectural and algorithmic innovations. By combining a frozen vision transformer backbone with a Laplacian residual branch, the model effectively captures both global semantic context and fine-grained textural artefacts pivotal for distinguishing morphing attacks. The introduction of residual statistic gated low-rank adapters (R-FLoRA) enables sample-adaptive modulation of the backbone’s feature extraction, allowing the model to emphasise morph-specific cues without overfitting to training set. Furthermore, the use of feature-wise fusion (Res-FiLM) ensures that crucial local evidence is preserved and efficiently integrated with transformer tokens. 

Existing methods based on diffusion probabilistic models~\cite{ivanovska2023mad_ddpm}, self-supervised and teacher–student frameworks~\cite{ivanovska2025selfmad}, and few-shot learning strategies~\cite{tapia2025single} exhibit notably higher error rates. This decreased performance may be due to their sensitivity to  domain variability and limited data efficiency in this challenging scenario. Similarly, foundation model-based approaches, such as ViT-SVM~\cite{zhang2024ViT}, and domain-adaptive frameworks such as MADation~\cite{caldeira2025madationfacemorphingattack}, do not generalise as effectively, potentially due to their reliance on feature regularities prevalent in the training data but not preserved in independent evaluation cohorts.

\begin{table*}
\caption{Quantitative results of S-MAD techniques when evaluated on MD\#4 database (Train:Val MD\#1:MD\#2). BSCER values are averaged over Morph Types (MTs) (mean ±95\% CI).}
\label{tab:ms40_feret_frgc_main}
\begin{tabular}{lcccc}
\toprule
Method & D-EER\% & BSCER@MACER=1\% & BSCER@MACER=5\% & BSCER@MACER=10\% \\
\midrule
\textbf{Proposed Method }& \textbf{15.31} $\pm$ \textbf{1.07} & \textbf{42.71} $\pm$ \textbf{11.13} & \textbf{29.32} $\pm$ \textbf{7.26} & \textbf{20.88} $\pm$ \textbf{3.01} \\
Diffusion probabilistic models~\cite{ivanovska2023mad_ddpm} & 56.83 $\pm$ 1.17 & 100.00 $\pm$ 0 & 99.80 $\pm$ 0 & 98.16 $\pm$ 0.17 \\
Self-Supervision~\cite{ivanovska2025selfmad} & 27.59 $\pm$ 1.23 & 69.82 $\pm$ 21.86 & 52.55 $\pm$ 25.10 & 48.80 $\pm$ 24.10 \\
Distil-MAD~\cite{RiaCVIP-2025} & 24.46 $\pm$ 0.84 & 80.52 $\pm$ 1.05 & 51.00 $\pm$ 2.16 & 36.75 $\pm$ 1.45 \\
Few-Shot MAD (MobileNet)~\cite{tapia2025single} & 37.16 $\pm$ 0.77 & 86.23 $\pm$ 5.19 & 76.56 $\pm$ 7.11 & 68.96 $\pm$ 9.35 \\
Few-Shot MAD (ResNet)~\cite{tapia2025single} & 38.54 $\pm$ 1.33 & 90.07 $\pm$ 6.26 & 80.24 $\pm$ 10.67 & 73.26 $\pm$ 13.34 \\
MADation \cite{caldeira2025madationfacemorphingattack} & 38.69 $\pm$ 9.76 & 61.99 $\pm$ 11.02 & 45.47 $\pm$ 11.02 & 39.96 $\pm$ 12.78 \\
ViTSVM~\cite{zhang2024ViT} & 19.95 $\pm$ 1.00 & 58.84 $\pm$ 15.78 & 45.04 $\pm$ 17.30 & 34.85 $\pm$ 17.04 \\
LocalMorph~\cite{qin2022face} & 32.26 $\pm$ 2.36 & 90.96 $\pm$ 18.88 & 49.39 $\pm$ 12.25 & 33.53 $\pm$ 12.79 \\
FS-SPN~\cite{zhang2018face} & 32.33 $\pm$ 4.58 & 86.74 $\pm$ 12.47 & 71.68 $\pm$ 3.74 & 63.25 $\pm$ 6.86 \\
\bottomrule
\end{tabular}
\end{table*}
Table~\ref{tab:ms40_feret_frgc_main} summarises the quantitative performance of the proposed and existing S-MAD methods on the FRILL dataset, including 95\% confidence intervals for each reported metric. 
The results in Table~\ref{tab:ms40_feret_frgc_main} show that the proposed S-MAD method provides the best detection accuracy on the MD\#4 database under cross-dataset evaluation (Train: MD\#1, Validation: MD\#2, Test: MD\#4), outperforming all state-of-the-art methods. With a D-EER of $15.31 \pm 1.07$ and notably lower BSCER values across all operating points, the proposed method exhibits robust generalisation to morphing artefacts even with heterogeneous training and evaluation domains. A key factor enabling this generalisation is the model’s effective exploitation of learned forensic priors during validation and its regularisation strategies, which discourage over-specialisation to any single database. By actively leveraging validation feedback and diverse morph types, the method avoids bias towards dataset-specific artefacts and demonstrates reliable discrimination across unseen samples. 

In contrast, baseline approaches~\cite{ivanovska2023mad_ddpm,ivanovska2025selfmad,RiaCVIP-2025,tapia2025single,caldeira2025madationfacemorphingattack,zhang2024ViT, qin2022face, zhang2018face} exhibit significantly higher error rates and greater susceptibility to performance degradation, likely reflecting their limited sensitivity to subtle cross-domain variations and their dependence on homogeneous training distributions.

\subsubsection{\textbf{Quantitative Results: train:MD\#2 dataset and Validation:MD\#1 dataset.}}

\begin{table*}
\caption{Quantitative results of S-MAD techniques when evaluated on MD\#3 database (Train:Val MD\#2:MD\#1).  BSCER values are averaged over Morph Types (MTs) (mean ±95\% CI).}
\label{tab:frill_frgc_feret_main}
\begin{tabular}{lcccc}
\toprule
Method & D-EER\% & BSCER@MACER=1\% & BSCER@MACER=5\% & BSCER@MACER=10\% \\
\midrule
\textbf{Proposed Method} & \textbf{4.00} $\pm$ \textbf{1.23} & \textbf{16.59} $\pm$ \textbf{13.05} & \textbf{4.61} $\pm$ \textbf{3.38} & \textbf{3.23} $\pm$ \textbf{0.54} \\
Diffusion probabilistic models~\cite{ivanovska2023mad_ddpm} & 66.67 $\pm$ 1.78 & 100.00 $\pm$ 0 & 100.00 $\pm$ 0 & 100.00 $\pm$ 0 \\
Self-Supervision~\cite{ivanovska2025selfmad} & 28.93 $\pm$ 1.94 & 73.81 $\pm$ 12.75 & 57.98 $\pm$ 13.59 & 49.86 $\pm$ 12.88 \\
Distil-MAD~\cite{RiaCVIP-2025} & 13.72 $\pm$ 1.03 & 66.67 $\pm$ 1.45 & 26.47 $\pm$ 2.13 & 15.69 $\pm$ 1.45 \\
Few-Shot MAD (MobileNet)~\cite{tapia2025single} & 33.60 $\pm$ 1.82 & 80.39 $\pm$ 9.66 & 69.89 $\pm$ 13.45 & 59.38 $\pm$ 15.27 \\
Few-Shot MAD (ResNet)~\cite{tapia2025single} & 48.85 $\pm$ 2.23 & 94.96 $\pm$ 0.56 & 92.30 $\pm$ 1.26 & 86.97 $\pm$ 3.64 \\
MADation \cite{caldeira2025madationfacemorphingattack} & 31.03 $\pm$ 2.22 & 80.25 $\pm$ 7.98 & 69.05 $\pm$ 10.36 & 57.42 $\pm$ 12.32 \\
ViTSVM~\cite{zhang2024ViT} & 42.73 $\pm$ 1.62 & 90.48 $\pm$ 2.80 & 78.43 $\pm$ 4.20 & 71.85 $\pm$ 5.60 \\
LocalMorph~\cite{qin2022face} & 27.45 $\pm$ 1.26 & 68.13 $\pm$ 7.28 & 59.12 $\pm$ 15.52 & 46.27 $\pm$ 8.17 \\
FS-SPN~\cite{zhang2018face} & 31.51 $\pm$ 2.58 & 94.11 $\pm$ 12.33 & 82.35 $\pm$ 8.78 & 60.78 $\pm$ 9.45 \\
\bottomrule
\end{tabular}
\end{table*}
Table~\ref{tab:frill_frgc_feret_main} summarises the quantitative performance of the proposed and existing S-MAD methods on the MD\#3 dataset, including 95\% confidence intervals for each reported metric. 
The results shown in Table~\ref{tab:frill_frgc_feret_main} provide compelling evidence of the proposed method's superior robustness and generalisation when subjected to a cross-dataset training protocol, with MD\#2 for training and MD\#1 for validation. The proposed method achieves the lowest D-EER and BSCER values amongst all evaluated techniques, indicating not only superior overall detection accuracy but also consistent performance at stringent operating points. This resilience is especially noteworthy given the inherent challenges posed by distinct acquisition settings, facial demographics, and morphing protocols present in the FRILL database.

Comparative state-of-the-art methods, including diffusion probabilistic models~\cite{ivanovska2023mad_ddpm}, self-supervision-based approaches~\cite{ivanovska2025selfmad}, Localised Features \cite{qin2022face}, Camera Noise \cite{zhang2018face}, teacher–student distillation~\cite{RiaCVIP-2025}, few-shot learning strategies~\cite{tapia2025single}, large-scale adaptive models~\cite{caldeira2025madationfacemorphingattack}, and transformer-based SVMs~\cite{zhang2024ViT}, all demonstrate elevated error rates and pronounced sensitivity to domain shift. Their degrading detection results across MACER thresholds highlight the limitations of methods relying predominantly on homogeneous training or implicit domain adaptation, which are often inadequate for morphing attack scenarios involving unseen subject matter and morph workflows.

In contrast, the proposed method’s validation-driven regularisation and capacity to incorporate feedback from diverse morph types during development underpin its ability to generalise effectively. These results confirm that robust cross-domain S-MAD performance demands both discriminative feature extraction and rigorous evaluation on heterogeneous validation sets, a criterion convincingly met by the proposed approach.

\begin{table*}
\caption{Quantitative results of S-MAD techniques when evaluated on MD\#4 database (Train:Val MD\#2:MD\#1). BSCER values are averaged over Morph Types (MTs) (mean ±95\% CI).}
\label{tab:ms40_frgc_feret_main}
\begin{tabular}{lcccc}
\toprule
Method & D-EER\% & BSCER@MACER=1\% & BSCER@MACER=5\% & BSCER@MACER=10\% \\
\midrule
\textbf{Proposed Method} & \textbf{20.23} $\pm$ \textbf{1.25} & \textbf{55.77} $\pm$ \textbf{16.70} & \textbf{40.22} $\pm$ \textbf{13.02} & \textbf{31.44} $\pm$ \textbf{11.85} \\
Diffusion probabilistic models~\cite{ivanovska2023mad_ddpm} & 57.64 $\pm$ 1.28 & 99.54 $\pm$ 0.06 & 99.11 $\pm$ 0.11 & 98.28 $\pm$ 0.14 \\
Self-Supervision~\cite{ivanovska2025selfmad} & 37.88 $\pm$ 1.27 & 78.51 $\pm$ 6.45 & 68.39 $\pm$ 8.00 & 61.65 $\pm$ 9.12 \\
Distil-MAD~\cite{RiaCVIP-2025} & 37.26 $\pm$ 1.03 & 86.35 $\pm$ 2.43 & 68.47 $\pm$ 4.23 & 56.43 $\pm$ 3.25 \\
Few-Shot MAD (MobileNet)~\cite{tapia2025single} & 37.16 $\pm$ 0.78 & 86.23 $\pm$ 5.22 & 76.56 $\pm$ 6.94 & 68.96 $\pm$ 9.35 \\
Few-Shot MAD (ResNet)~\cite{tapia2025single} & 38.54 $\pm$ 1.31 & 90.07 $\pm$ 5.82 & 80.24 $\pm$ 10.44 & 73.26 $\pm$ 12.79 \\
MADation \cite{caldeira2025madationfacemorphingattack} & 28.28 $\pm$ 1.27 & 69.79 $\pm$ 10.79 & 56.34 $\pm$ 10.30 & 45.96 $\pm$ 9.58 \\
ViTSVM~\cite{zhang2024ViT} & 34.90 $\pm$ 1.51 & 93.20 $\pm$ 1.72 & 78.66 $\pm$ 4.30 & 66.87 $\pm$ 5.22 \\
LocalMorph~\cite{qin2022face} & 21.68 $\pm$ 3.75 & 87.34 $\pm$ 5.66 & 62.65 $\pm$ 14.25 & 56.62 $\pm$ 11.21 \\
FS-SPN~\cite{zhang2018face} & 32.32 $\pm$ 4.78 & 86.74 $\pm$ 8.55 & 71.68 $\pm$ 12.45 & 63.25 $\pm$ 14.11 \\
\bottomrule
\end{tabular}
\end{table*}
Table~\ref{tab:ms40_frgc_feret_main} summarises the quantitative performance of the proposed and existing S-MAD methods on the MD\#3 dataset, including 95\% confidence intervals for each reported metric. 
Table~\ref{tab:ms40_frgc_feret_main} presents the detection results for the MD\#4 dataset using a cross-dataset protocol (MD\#2 for training, MD\#1 for validation). The proposed method emerges as the most robust among all evaluated approaches, achieving the lowest D-EER and BSCER values despite the considerable distributional gap between the training/validation sources and the MD\#4 evaluation cohort.

The observed robustness evidences the proposed method capacity for meaningful cross-domain generalisation, successfully adapting to variations in acquisition protocols, demographic composition, and morph generation processes in MD\#4. Strict validation with MD\#1 ensures comprehensive tuning, enabling reliable discrimination in operationally realistic and unseen environments.

In contrast, existing techniques including diffusion probabilistic models~\cite{ivanovska2023mad_ddpm}, self-supervised learning~\cite{ivanovska2025selfmad}, LocalMorph~\cite{qin2022face}, FS-SPN~\cite{zhang2018face}, distillation-based architectures~\cite{RiaCVIP-2025}, few-shot learning~\cite{tapia2025single}, domain adaptation frameworks~\cite{caldeira2025madationfacemorphingattack}, and transformer-SVMs~\cite{zhang2024ViT} indicates a pronounced decreases in accuracy and high false positive rates. Their limited ability to adjust to the specific characteristics of the MD\#4 database reflects insufficient domain adaptation and restricted robustness to cross-dataset morphing variations.

These results highlight the importance of evaluation protocols incorporating heterogeneous validation and test sets and confirm that only methods with genuinely transferable forensic priors and careful validation-driven regularisation are suited for deployment in MAD across diverse biometric sources.  Detection Error Trade-off (DET) curve visualisations for all evaluation protocols and datasets are available in the supplementary material.
\section{Ablation Study}
\label{sec:Ablation}

\begin{table*}[t]
\centering
\setlength{\tabcolsep}{4pt} 
\small                      
\resizebox{\textwidth}{!}
{%
\begin{tabular}{lccccccc|rrrr}
\toprule
\textbf{Case} & \textbf{R-FLoRA} & \textbf{Q} & \textbf{V} & \textbf{k} & \textbf{Residual} & \textbf{Fusion} & \textbf{RCA} 
& \textbf{D-EER\%} $\downarrow$ & \textbf{BSCER@1\%} $\downarrow$ & \textbf{@5\%} $\downarrow$ & \textbf{@10\%} $\downarrow$ \\
\midrule
Fusion-only (baseline w/o adapters)          & \xmark & \xmark & \xmark & 0  & \cmark & Res-FiLM  & \cmark & 6.12 & 11.76 & 6.86 & 5.88 \\
k=2 (Q+V)                                   & \cmark & \cmark & \cmark & 2  & \cmark & Res-FiLM  & \cmark & 6.05 & 17.83 & 11.45 & 7.86 \\
\textbf{k=3 (Q+V)  (Proposed)}              & \cmark & \cmark & \cmark & \textbf{3} & \cmark & Res-FiLM  & \cmark & \textbf{4.94} & \textbf{18.63} & \textbf{6.02} & \textbf{3.08} \\
k=6 (Q+V)                                   & \cmark & \cmark & \cmark & 6  & \cmark & Res-FiLM  & \cmark & 7.54 & 33.33 & 9.80 & 6.86 \\
k=12 (Q+V)                                  & \cmark & \cmark & \cmark & 12 & \cmark & Res-FiLM  & \cmark & 9.75 & 33.33 & 15.80 & 9.80 \\
Q-only (k=3)                                & \cmark & \cmark & \xmark & 3  & \cmark & Res-FiLM  & \cmark & 7.28 & 19.87 & 8.68 & 5.90 \\
V-only (k=3)                                & \cmark & \xmark & \cmark & 3  & \cmark & Res-FiLM  & \cmark & 6.18 & 17.27 & 7.42 & 5.20 \\
RCA off (k=3)                               & \cmark & \cmark & \cmark & 3  & \cmark & Res-FiLM  & \xmark & 6.86 & 30.39 & 6.86 & 4.90 \\
No residual branch                          & \xmark & \xmark & \xmark & 0  & \xmark & ---       & \xmark & 15.78 & 24.50 & 24.50 & 24.50 \\
\bottomrule
\end{tabular}
}
\caption{Ablation on key components of the proposed S-MAD approach. 
Training is performed on the MD\#2 dataset, and ablations are conducted on the MD\#1 \textit{development} set; MD\#3  remain strictly unseen for evaluation. 
\cmark/\xmark\ denote presence/absence of each component. 
Metrics are reported as D-EER (↓ better) and BSCER at fixed MACER levels (1\%, 5\%, 10\%). 
Boldface indicates the proposed configuration.}
\label{tab:ablation}
\end{table*}

Table \ref{tab:ablation} analyses the contribution of each major component within the proposed S-MAD framework. The baseline configuration using only the residual fusion (Res-FiLM) without any R-FLoRA adaptation achieves a D-EER of 6.12\%. Introducing the R-FLoRA adapters progressively enhances discriminability when a limited number of layers are adapted. The optimal performance is observed for k = 3 layers, with adapters attached to both query and value projections, yielding the lowest D-EER (4.94\%) and consistently improved BSCER across all operating points. This suggests that a shallow and jointly conditioned adaptation is sufficient to capture morph-specific residual cues, whereas deeper injection ($k \geq 6$) introduces redundant modulation and potential over-adaptation, degrading generalisation.

Disabling the residual branch causes a sharp performance drop (D-EER = 15.78\%), confirming that the Laplacian-based residual cue is the primary discriminative signal for morphing artefacts. Removing the RCA alignment loss also leads to higher D-EER (6.86\%) and elevated BSCER, indicating that RCA effectively regularises the token-space consistency between the original and fused representations. The comparison between Q-only and V-only variants further shows that modulation through both attention query and value channels contributes complementary information: query-only adaptation improves focus on discriminative regions, while value-only adaptation enhances local evidence integration; their combination provides the most balanced result.

Overall, the ablation results demonstrate that the proposed configuration (k = 3, Q+V, Res-FiLM, RCA enabled) offers the best trade-off between adaptation depth and stability, achieving robust detection without over-specialisation. This justifies its selection for all subsequent experiments on the unseen MD\#3 evaluation datasets.

\subsection{Robustness to Controlled Image Degradations}
\label{sec:e8_robustness}

This experiment evaluates the robustness of morphing attack detection models under controlled image degradations without any retraining or domain adaptation. All models are trained on MD\#2 and validated on the MD\#1 following the standard experimental protocol and are subsequently evaluated on \emph{unseen} test domains, namely MD\#3 dataset. This setting explicitly assesses cross-domain generalisation under realistic image quality variations.

\begin{figure}[!t]
\centering
\includegraphics[width=\linewidth]{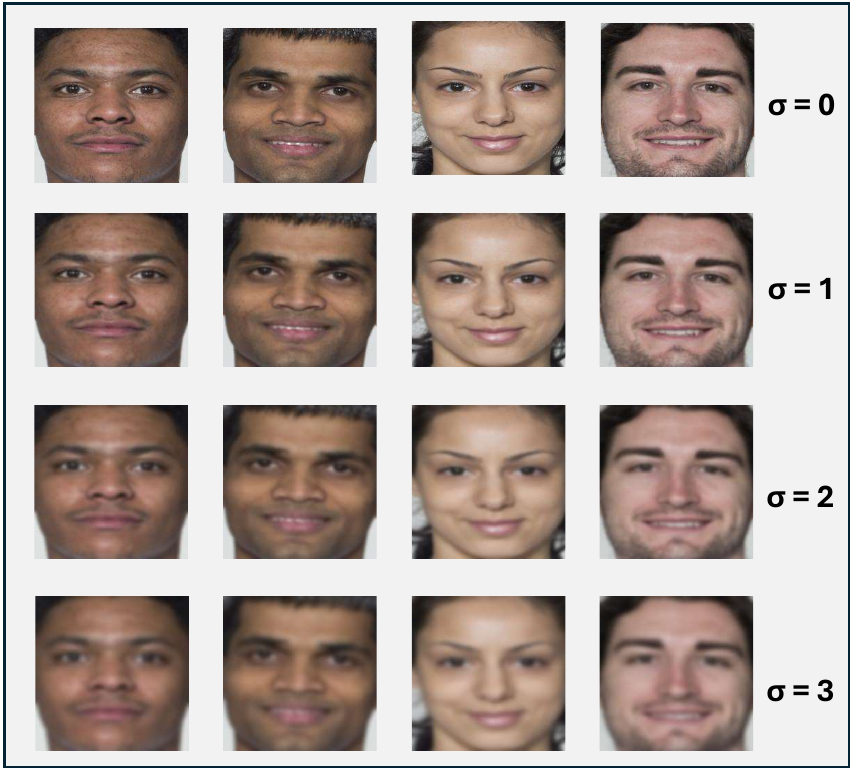}
\caption{Examples of Gaussian blur degradation applied at test time with increasing standard deviation $\sigma \in \{0,1,2,3\}$.}
\label{fig:e8_blur}
\end{figure}

\begin{figure}[!t]
\centering
\includegraphics[width=\linewidth]{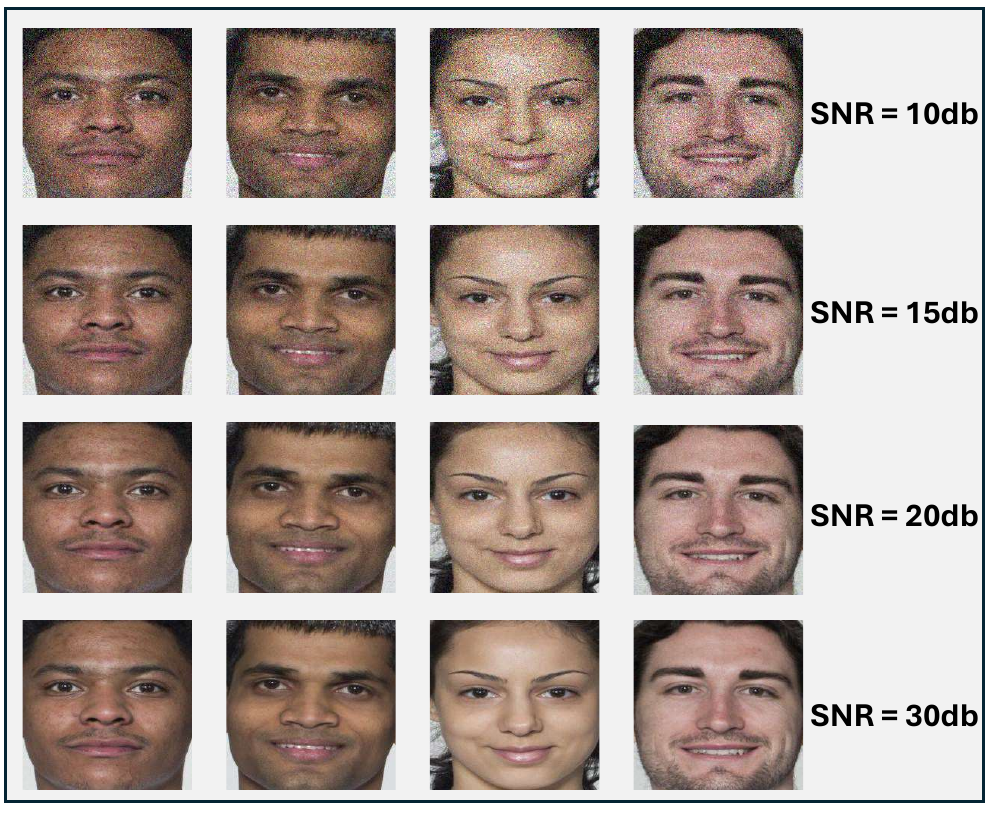}
\caption{Examples of additive Gaussian noise applied at test time with decreasing signal-to-noise ratio levels (30~dB to 10~dB).}
\label{fig:e8_noise}
\end{figure}

\begin{figure}[!t]
\centering
\includegraphics[width=\linewidth]{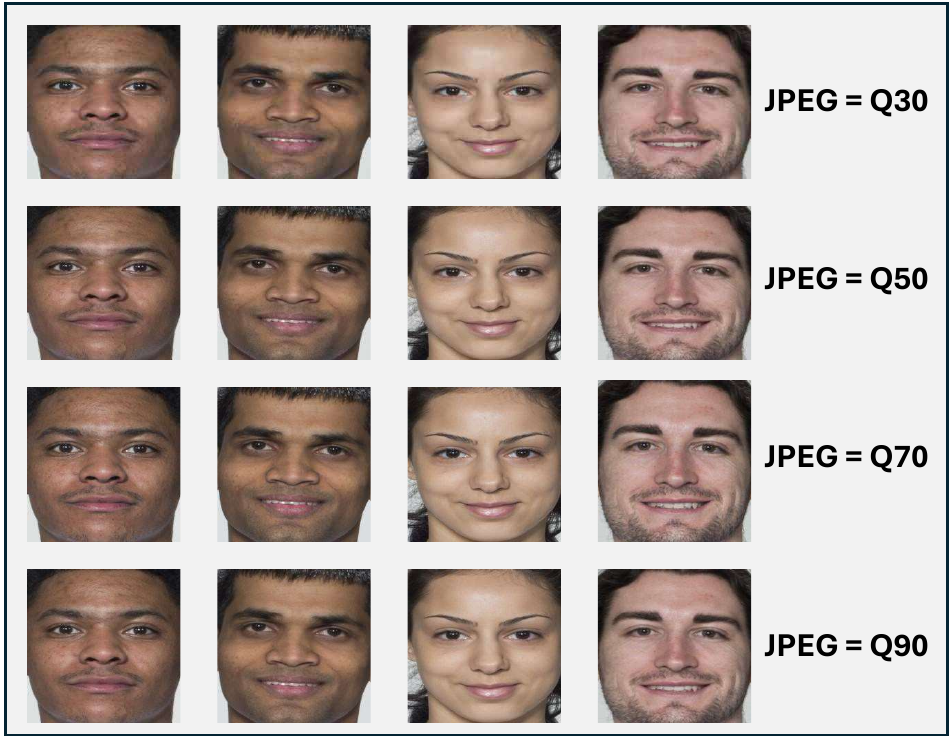}
\caption{Examples of JPEG compression applied at test time with decreasing quality factors $q \in \{90,70,50,30\}$.}
\label{fig:e8_jpeg}
\end{figure}

Three degradation types are considered: Gaussian blur, additive Gaussian noise, and JPEG compression. The degradations are applied only at test time, ensuring that no degraded samples are observed during training. Gaussian blur is introduced with standard deviation values $\sigma \in \{0,1,2,3\}$. Additive Gaussian noise is applied at signal-to-noise ratio levels of 30~dB, 20~dB, 15~dB, and 10~dB. JPEG compression is simulated using quality factors $q \in \{90,70,50,30\}$. Representative examples of each degradation level are illustrated in Fig.~\ref{fig:e8_blur}, Fig.~\ref{fig:e8_noise}, and Fig.~\ref{fig:e8_jpeg}.

\begin{table}[!t]
\centering
\caption{Robustness evaluation under blur, additive noise, and JPEG compression on the MD\#3  test set: mean EER (\%) and 95\% confidence intervals, where statistics are computed across the seven MD\#3  morphing attack generation methods.}
\label{tab:robustness_proposed_eer_ci}
\setlength{\tabcolsep}{6pt}
\resizebox{\linewidth}{!}{
\begin{tabular}{lc}
\hline
\textbf{Condition} & \textbf{Proposed Method: Mean EER [CI 95\%]} \\
\hline
Clean                & 4.00 [2.77, 5.22] \\
Blur ($\sigma=0$)    & 4.00 [2.77, 5.22] \\
Blur ($\sigma=1$)    & 8.74 [7.27, 10.21] \\
Blur ($\sigma=2$)    & 22.70 [19.87, 25.52] \\
Blur ($\sigma=3$)    & 27.25 [23.13, 31.36] \\
Noise (SNR = 5 dB)   & 28.53 [25.12, 31.94] \\
Noise (SNR = 10 dB)  & 20.49 [18.20, 22.77] \\
Noise (SNR = 20 dB)  & 11.58 [7.32, 15.84] \\
Noise (SNR = 30 dB)  & 8.93 [5.60, 12.25] \\
JPEG (Q = 30)        & 12.77 [12.49, 13.04] \\
JPEG (Q = 50)        & 8.07 [6.64, 9.50] \\
JPEG (Q = 70)        & 4.98 [2.82, 7.13] \\
JPEG (Q = 90)        & 3.94 [2.70, 5.18] \\
\hline
\end{tabular}
}
\end{table}

Quantitative robustness results on the MD\#3 digital test set are summarised in Table~\ref{tab:robustness_proposed_eer_ci}. The reported metrics correspond to the mean equal error rate (EER) and 95\% confidence intervals computed across the seven MD\#3 morphing attack generation methods. By aggregating performance statistics over morphing types, the analysis captures variability arising from different morphing processes rather than fluctuations due to individual test samples.

Overall, the proposed method exhibits stable and predictable behaviour across a wide range of realistic image degradations. Error rates increase gradually as degradation severity intensifies, while the associated confidence intervals remain relatively tight for most conditions. This indicates consistent performance across morphing attack types and limited sensitivity to any particular attack generation mechanism.

Under Gaussian blur, the proposed method shows a smooth and monotonic increase in EER as the blur parameter $\sigma$ increases. Performance remains close to the clean condition for mild blur and degrades in a controlled manner for stronger blur levels. This behaviour suggests that the residual modelling strategy employed by the proposed approach is able to preserve discriminative information even when fine-grained spatial detail is progressively attenuated by low-pass filtering.

For additive noise, the proposed approach maintains low error rates at moderate signal-to-noise ratios (30~dB and 20~dB), with more pronounced degradation observed only at lower SNR values. Importantly, no abrupt performance collapse is observed, even under severe noise conditions, indicating robustness to stochastic perturbations commonly encountered in unconstrained image acquisition scenarios.

JPEG compression has a comparatively limited impact on performance, particularly at quality factors of 90 and 70, where the EER remains close to the clean baseline. Even at lower quality settings, the increase in error remains gradual. This suggests that the learned representation is not overly dependent on compression-sensitive high-frequency artefacts, but instead leverages more stable structural and residual cues that are preserved under lossy compression.

Overall, these results demonstrate that the proposed method generalises effectively under realistic image degradations without requiring explicit exposure to degraded data during training. The observed robustness can therefore be attributed to the intrinsic design of the residual representation and adaptation mechanism, rather than to degradation-specific optimisation. This property is particularly important for forensic and biometric applications, where variations in image quality are unavoidable and cannot be fully anticipated at training time.

\section{Limitations and Future Work}
\label{sec:Lim}

Although the proposed framework shows strong performance across four ICAO compliant databases and several morph generation techniques, a number of limitations should be considered. First, the evaluation has focused on controlled digital images, and the behaviour of the method under severe quality degradation has not yet been established. Examples include low resolution surveillance imagery, strong compression and other conditions where high frequency detail may be unreliable. Second, while the model can handle a wide range of known morphing procedures, its response to deliberately engineered attacks that are optimised to suppress both Laplacian cues and transformer based cues remains uncertain. Understanding these cases requires further forensic study. Third, the current design is aimed at digital photograph submission workflows. Situations that involve print and scan, recapture or other acquisition processes may introduce distortions that fall outside the conditions explored in this study, and these settings would require additional calibration of the proposed method.

Future work will examine the robustness of the method against more advanced morph generation strategies and assess its performance across a broader set of sensing conditions, including mobile and embedded systems. It would also be useful to study workflow dependent distortions such as print and scan or recapture artefacts. These effects are relevant in some document issuance procedures but are difficult to evaluate in a consistent way because the outcome depends strongly on the device and its settings. Developing principled ways to model or simulate such distortions would provide a clearer understanding of the limits of the method under physical acquisition. Further improvement may be possible by incorporating additional sources of information, for example temporal consistency, contextual metadata or sensor related characteristics, in order to support reliable decisions in complex operational environments. Finally, detailed interpretability studies based on visual attribution methods such as Grad CAM or token relevance mapping may offer further insight into the image regions used by the model and may support clearer forensic interpretation in practice.

\section{Conclusion}
\label{sec:conc}

This paper has presented a new framework for S-MAD that integrates Laplacian residual statistics with frozen foundation-scale vision transformer backbones. The proposed architecture employs residual-statistic-gated low-rank adapters and content-adaptive fusion layers to achieve discriminative and interpretable representations of morphing artefacts. Extensive cross-dataset experiments on four ICAO-compliant databases covering a wide range of morph generation techniques demonstrate the framework’s effectiveness and generalisation capability.
Quantitative evaluations show consistent and significant improvements over nine recent state-of-the-art methods, confirming the robustness and transferability of the proposed approach under previously unseen and operationally realistic conditions. Ablation analyses further establish the importance of each architectural component, emphasising the complementary role of local residual cues and adaptive backbone modulation in achieving reliable detection.
Overall, this work advances the state of research in morphing attack detection by providing a technically rigorous, computationally efficient, and practically deployable solution for secure identity verification and border control systems.



{
\small{
\bibliographystyle{IEEEtran}
\bibliography{FaceMorph_Refereces}
}
}
\end{document}